\documentclass[11pt,a4paper]{article}
\usepackage[hyperref]{acl2018}
\usepackage{times}
\usepackage{latexsym}
\usepackage[shortlabels]{enumitem}
                    \setlist[enumerate, 1]{1\textsuperscript{o}}
\usepackage{dsfont}

\usepackage{bm}
\usepackage{url}
\usepackage{amsmath}
\usepackage{amssymb}
\usepackage{graphicx} 
\usepackage{diagbox}
\usepackage{caption}
\usepackage{booktabs} 

\newcommand{\vect}[1]{\bm{#1}}
\newcommand{\mat}[1]{\bf{#1}}
\newcommand{\RR}{\mathbb{R}}

\aclfinalcopy 

\title{Neural Natural Language Inference Models Enhanced with \\ External Knowledge}

\author{
Qian Chen \\
University of Science and \\ 
Technology of China\\
\tt{cq1231@mail.ustc.edu.cn} \\\And
Xiaodan Zhu \\
ECE, Queen's University \\
\texttt{xiaodan.zhu@queensu.ca} \\\AND
Zhen-Hua Ling \\
University of Science and \\
Technology of China\\
\tt{zhling@ustc.edu.cn} \\\And
Diana Inkpen \\
University of Ottawa\\
\tt{diana@site.uottawa.ca} \\\AND
Si Wei \\
iFLYTEK Research\\
\tt{siwei@iflytek.com} \\
}
\date{}

\begin{document}
\maketitle
\begin{abstract}
Modeling natural language inference is a very challenging task. With the availability of large annotated data, it has recently become feasible to train complex models such as neural-network-based inference models, which have shown to achieve the state-of-the-art performance.
Although there exist relatively large annotated data, can machines learn all knowledge needed to perform natural language inference (NLI) from these data? If not, how can neural-network-based NLI models benefit from external knowledge and how to build NLI models to leverage it?
In this paper, we enrich the state-of-the-art neural natural language inference models with external knowledge. We demonstrate that the proposed models improve neural NLI models to achieve the state-of-the-art performance on the SNLI and MultiNLI datasets.

\end{abstract}

\section{Introduction}

Reasoning and inference are central to both human and artificial intelligence. Natural language inference (NLI), also known as recognizing textual entailment (RTE), is an important NLP problem concerned with determining inferential relationship (e.g., entailment, contradiction, or neutral) between a premise~\textit{p} and a hypothesis~\textit{h}. In general, modeling informal inference in language is a very challenging and basic problem towards achieving true natural language understanding.

In the last several years, larger annotated datasets were made available, e.g., the SNLI~\citep{DBLP:conf/emnlp/BowmanAPM15} and MultiNLI datasets~\citep{DBLP:journals/corr/WilliamsNB17}, which made it feasible to train rather complicated neural-network-based models that fit a large set of parameters to better model NLI. Such models have shown to achieve the state-of-the-art performance~\citep{DBLP:conf/emnlp/BowmanAPM15,DBLP:conf/acl/BowmanGRGMP16,DBLP:conf/eacl/YuM17,DBLP:conf/emnlp/ParikhT0U16,DBLP:conf/coling/ShaCSL16,DBLP:conf/acl/ChenZLWJI17,DBLP:conf/repeval/ChenZLWJI17,DBLP:journals/corr/abs-1801-00102}. 

While neural networks have been shown to be very effective in modeling NLI with large training data, they have often focused on end-to-end training by assuming that  all inference knowledge is learnable from the provided training data. In this paper, we relax this assumption and explore whether external knowledge can further help NLI. Consider an example:
\begin{itemize}
\itemsep 0em
\item \textit{p}: A lady standing in a \textit{wheat} field.	
\item \textit{h}: A person standing in a \textit{corn} field. 
\end{itemize}
In this simplified example, when computers are asked to predict the relation between these two sentences and if training data do not provide the knowledge of relationship between ``wheat'' and ``corn'' (e.g., if one of the two words does not appear in the training data or they are not paired in any premise-hypothesis pairs), it will be hard for computers to correctly recognize that the premise contradicts the hypothesis. 

In general, although in many tasks learning \textit{tabula rasa} achieved state-of-the-art performance, we believe complicated NLP problems such as NLI could benefit from leveraging knowledge accumulated by humans, particularly in a foreseeable future when machines are unable to learn it by themselves. 

In this paper we enrich neural-network-based NLI models with external knowledge in co-attention, local inference collection, and inference composition components. We show the proposed model improves the state-of-the-art NLI models to achieve better performances on the SNLI and MultiNLI datasets. The advantage of using external knowledge is more significant when the size of training data is restricted, suggesting that if more knowledge can be obtained, it may bring more benefit. In addition to attaining the state-of-the-art performance, we are also interested in understanding how external knowledge contributes to the major components of typical neural-network-based NLI models. 

\section{Related Work}
\label{sec:related}
Early research on natural language inference and recognizing textual entailment has been performed on relatively small datasets (refer to~\citet{MacCartneyThesis} for a good literature survey), which includes a large bulk of contributions made under the name of RTE, such as~\citep{Dagan2005ThePR,Iftene:W07-1421}, among many others.   

More recently the availability of much larger annotated data, e.g., SNLI ~\citep{DBLP:conf/emnlp/BowmanAPM15} and MultiNLI~\citep{DBLP:journals/corr/WilliamsNB17}, has made it possible to train more complex models. These models mainly fall into two types of approaches: sentence-encoding-based models and models using also inter-sentence attention. Sentence-encoding-based models use Siamese architecture~\citep{DBLP:conf/nips/BromleyGLSS93}. The parameter-tied neural networks are applied to encode both the premise and the hypothesis. Then a neural network classifier is applied to decide relationship between the two sentences. Different neural networks have been utilized for sentence encoding, such as LSTM~\citep{DBLP:conf/emnlp/BowmanAPM15}, GRU~\citep{DBLP:journals/corr/VendrovKFU15}, CNN~\citep{DBLP:conf/acl/MouMLX0YJ16}, BiLSTM and its variants~\citep{DBLP:journals/corr/LiuSLW16,DBLP:journals/corr/LinFSYXZB17,DBLP:conf/repeval/ChenZLWJI17,DBLP:conf/repeval/NieB17}, self-attention network~\citep{DBLP:journals/corr/abs-1709-04696,DBLP:journals/corr/abs-1801-10296}, and more complicated neural networks~\citep{DBLP:conf/acl/BowmanGRGMP16,DBLP:conf/eacl/YuM17a,DBLP:conf/eacl/YuM17,DBLP:journals/corr/ChoiYL17}. 
Sentence-encoding-based models transform sentences into fixed-length vector representations, which may help a wide range of tasks~\citep{DBLP:conf/emnlp/ConneauKSBB17}. 

The second set of models use inter-sentence attention~\citep{DBLP:journals/corr/RocktaschelGHKB15,DBLP:conf/naacl/WangJ16,DBLP:conf/emnlp/0001DL16,DBLP:conf/emnlp/ParikhT0U16,DBLP:conf/acl/ChenZLWJI17}. Among them,~\citet{DBLP:journals/corr/RocktaschelGHKB15} were among the first to propose neural attention-based models for NLI.~\citet{DBLP:conf/acl/ChenZLWJI17} proposed an enhanced sequential inference model (ESIM), which is one of the best models so far and is used as one of our baselines in this paper.

In this paper we enrich neural-network-based NLI models with external knowledge. Unlike early work on NLI~\citep{Jijkoun2005RecognizingTE, DBLP:conf/emnlp/MacCartneyGM08,MacCartneyThesis} that explores external knowledge in conventional NLI models on relatively small NLI datasets, we aim to merge the advantage of powerful modeling ability of neural networks with extra external inference knowledge. We show that the proposed model improves the state-of-the-art neural NLI models to achieve better performances on the SNLI and MultiNLI datasets. The advantage of using external knowledge is more significant when the size of training data is restricted, suggesting that if more knowledge can be obtained, it may have more benefit. In addition to attaining the state-of-the-art performance, we are also interested in understanding how external knowledge affect major components of neural-network-based NLI models.

In general, external knowledge has shown to be effective in neural networks for other NLP tasks, including 
word embedding~\citep{DBLP:conf/acl/ChenLCCWJZ15,DBLP:conf/naacl/FaruquiDJDHS15,DBLP:conf/acl/Liu0WLH15,DBLP:journals/tacl/WietingBGL15,DBLP:journals/corr/MrksicVSLRGKY17}, machine translation~\citep{DBLP:conf/acl/ShiLRFLZSW16,DBLP:conf/apsipa/ZhangMWH17}, language modeling~\citep{DBLP:journals/corr/AhnCPB16}, and dialogue systems~\citep{DBLP:journals/corr/ChenHTCGD16}. 

\section{Neural-Network-Based NLI Models with External Knowledge}

In this section we propose neural-network-based NLI models to incorporate external inference knowledge, which, as we will show later in Section \ref{sec:res}, achieve the state-of-the-art performance. In addition to attaining the leading performance we are also interested in investigating the effects of external knowledge on major components of  neural-network-based NLI modeling. 

Figure \ref{fig:model} shows a high-level general view of the proposed framework. While specific NLI systems vary in their implementation, typical state-of-the-art NLI models contain the main components (or equivalents) of representing premise and hypothesis sentences, collecting local (e.g., lexical) inference information, and aggregating and composing local information to make the global decision at the sentence level. We incorporate and investigate external knowledge accordingly in these major NLI components: computing co-attention, collecting local inference information, and composing inference to make final decision.

\subsection{External Knowledge}
\label{sec:knowledge}
As discussed above, although there exist relatively large annotated data for NLI, can machines learn all inference knowledge needed to perform NLI from the data? If not, how can neural network-based NLI models benefit from external knowledge and how to build NLI models to leverage it?

We study the incorporation of external, inference-related knowledge in major components of neural networks for natural language inference. For example, intuitively knowledge about \textit{synonymy}, \textit{antonymy}, \textit{hypernymy} and \textit{hyponymy} between given words may help model soft-alignment between premises and hypotheses; knowledge about \textit{hypernymy} and \textit{hyponymy} may help capture entailment; knowledge about \textit{antonymy} and \textit{co-hyponyms} (words sharing the same hypernym) may benefit the modeling of contradiction. 

\begin{figure*}[!htb]
	\centering
	\includegraphics[width=1\linewidth]{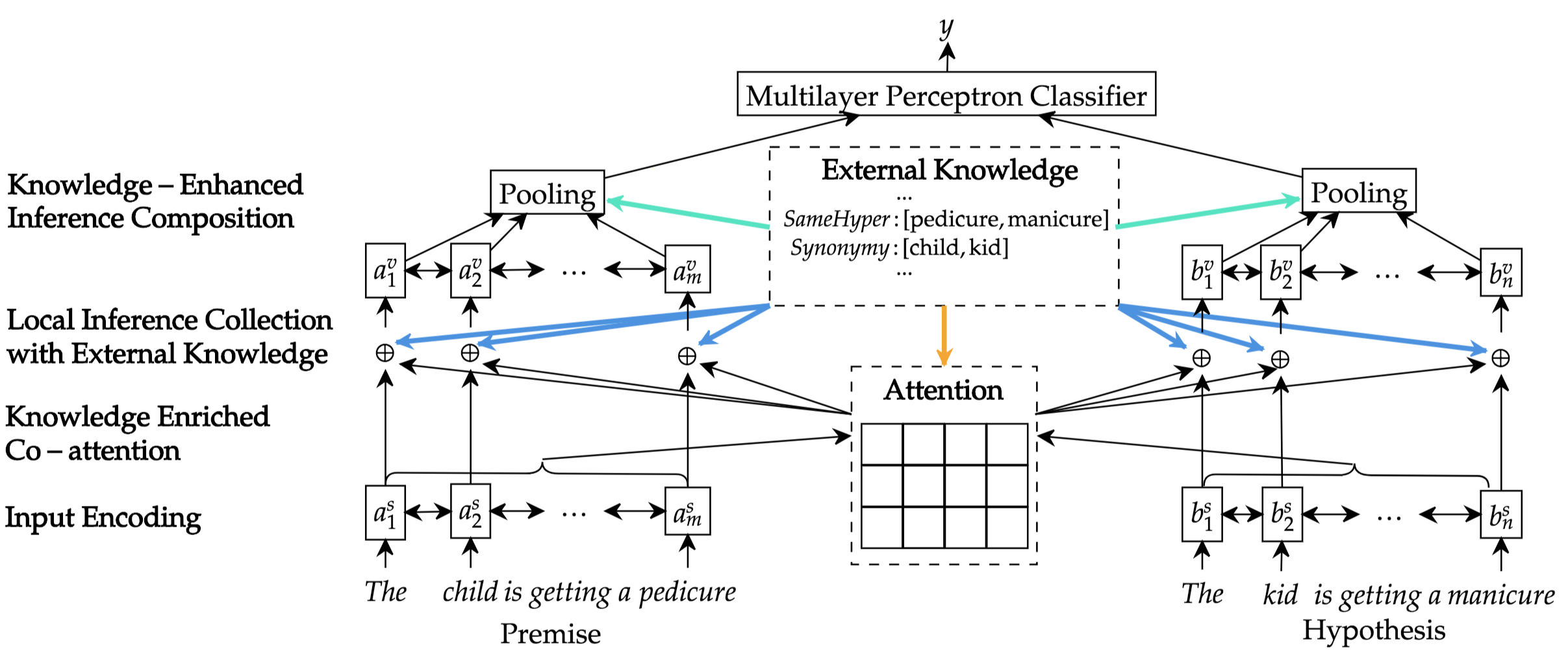}
	\caption{A high-level view of neural-network-based NLI models enriched with external knowledge in co-attention, local inference collection, and inference composition.}
	\label{fig:model}
\end{figure*}

In this section, we discuss the incorporation of basic, lexical-level semantic knowledge into neural NLI components. Specifically, we consider external lexical-level inference knowledge between word $w_i$ and $w_j$, which is represented as a vector ${\vect r}_{ij}$ and is incorporated into three specific components shown in Figure~\ref{fig:model}. We will discuss the details of how ${\vect r}_{ij}$ is constructed later in the experiment setup section (Section~\ref{sec:setup}) but instead focus on the proposed model in this section. Note that while we study lexical-level inference knowledge in the paper, if inference knowledge about larger pieces of text pairs (e.g., inference relations between phrases) are available, the proposed model can be easily extended to handle that. In this paper, we instead let the NLI models to compose lexical-level knowledge to obtain inference relations between larger pieces of texts. 

\subsection{Encoding Premise and Hypothesis}

Same as much previous work~\citep{DBLP:conf/acl/ChenZLWJI17,DBLP:conf/repeval/ChenZLWJI17}, we encode the premise and the hypothesis with bidirectional LSTMs (BiLSTMs). The premise is represented as  $\vect{a} = (a_1,\dots,a_m)$ and the hypothesis is $\vect{b}=(b_1,\dots,b_{n})$, where $m$ and $n$ are the lengths of the sentences. Then $\vect{a}$ and $\vect{b}$ are embedded into $d_e$-dimensional vectors $[{\mat E}(a_1),\dots,{\mat E}(a_m)]$ and $[{\mat E}(b_1),\dots,{\mat E}(b_n)]$ using the embedding matrix ${\mat E} \in \RR^{d_e \times |V|}$, where $|V|$ is the vocabulary size and ${\mat E}$ can be initialized with the pre-trained word embedding. To represent words in its context, the premise and the hypothesis are fed into BiLSTM encoders~\citep{DBLP:journals/neco/HochreiterS97} to obtain context-dependent hidden states ${\vect a}^s$ and ${\vect b}^s$: 
\begin{align}
{\vect a}^s_i &=\mathrm{Encoder}({\mat E}(\vect{a}),i) \,,\\
{\vect b}^s_j &=\mathrm{Encoder}({\mat E}(\vect{b}),j) \,.
\end{align}
where $i$ and $j$ indicate the $i$-th word in the premise and the $j$-th word in the hypothesis, respectively.

\subsection{Knowledge-Enriched Co-Attention}
As discussed above, soft-alignment of word pairs between the premise and the hypothesis may benefit from knowledge-enriched co-attention mechanism. Given the relation features ${\vect r}_{ij} \in \RR^{d_r}$ between the premise's $i$-th word and the hypothesis's $j$-th word derived from the external knowledge, the co-attention is calculated as:
\begin{equation}
e_{ij} = ({\vect a}^s_i)^\mathrm{T} {\vect b}^s_j + F({\vect r}_{ij}) \,.
\label{eq:eij}
\end{equation}
The function $F$ can be any non-linear or linear functions. In this paper, we use $F({\vect r}_{ij}) = \lambda \mathds{1}({\vect r}_{ij})$, where $\lambda$ is a hyper-parameter tuned on the development set and $\mathds{1}$ is the indication function as follows:
\begin{align}
\mathds{1}({\vect r}_{ij}) =
  \begin{cases}
    1  & \quad \text{if } {\vect r}_{ij} \text{ is not a zero vector} \,;\\
    0  & \quad \text{if } {\vect r}_{ij} \text{ is a zero vector} \,.
  \end{cases}
\end{align}
Intuitively, word pairs with semantic relationship, e.g., synonymy, antonymy, hypernymy, hyponymy and co-hyponyms, are probably aligned together. We will discuss how we construct external knowledge later in Section \ref{sec:setup}. We have also tried a two-layer MLP as a universal function approximator in function $F$ to learn the underlying combination function but did not observe further improvement over the best performance we obtained on the development datasets. 

Soft-alignment is determined by the co-attention matrix ${\mat e} \in \RR^{m \times n}$ computed in Equation~(\ref{eq:eij}), which is used to obtain the local relevance between the premise and the hypothesis. For the hidden state of the $i$-th word in the premise, i.e., ${\vect a}^s_i$ (already encoding the word itself and its context), the relevant semantics in the hypothesis is identified into a context vector ${\vect a}^c_i$ using $e_{ij}$, more specifically with Equation~(\ref{eq:a_c}).
\begin{align}
\label{eq:a_c}
\alpha_{ij} & = \frac{\exp(e_{ij})}{\sum_{k=1}^{n}\exp(e_{ik})} \,, ~{\vect a}^c_i =\sum_{j=1}^{n}\alpha_{ij} {\vect b}^s_j \,,\\
\label{eq:b_c}
\beta_{ij} & = \frac{\exp(e_{ij})}{\sum_{k=1}^{m}\exp(e_{kj})} \,, ~{\vect b}^c_j =\sum_{i=1}^{m}\beta_{ij} {\vect a}^s_i \,,
\end{align}
\noindent where ${\vect \alpha} \in \RR^{m \times n}$ and ${\vect \beta} \in \RR^{m \times n}$ are the normalized attention weight matrices with respect to the $2$-axis and $1$-axis. The same calculation is performed for each word in the hypothesis, i.e., ${\vect b}^s_j$, with Equation~(\ref{eq:b_c}) to obtain the context vector ${\vect b}^c_j$. 

\subsection{Local Inference Collection with External Knowledge}

By way of comparing the inference-related semantic relation between ${\vect a}_i^s$ (individual word representation in premise) and ${\vect a}_i^c$ (context representation from hypothesis which is align to word ${\vect a}^s_i$), we can model local inference (i.e., word-level inference) between aligned word pairs. Intuitively, for example, knowledge about hypernymy or hyponymy may help model entailment and knowledge about antonymy and co-hyponyms may help model contradiction. Through comparing ${\vect a}^s_i$ and ${\vect a}^c_i$, in addition to their relation from external knowledge, we can obtain word-level inference information for each word. The same calculation is performed for ${\vect b}_j^s$ and ${\vect b}_j^c$. Thus, we collect knowledge-enriched local inference information:
{\fontsize{10pt}{1.0cm}
\begin{align}
\label{eq:infer1}
{\vect a}^m_i &= G([{\vect a}^s_i;{\vect a}^c_i;{\vect a}^s_i - {\vect a}^c_i;{\vect a}^s_i \circ {\vect a}^c_i;\sum_{j=1}^{n} \alpha_{ij} {\vect r}_{ij}]) \,,\\
\label{eq:infer2}
{\vect b}^m_j &= G([{\vect b}^s_j,{\vect b}^c_j;{\vect b}^s_j - {\vect b}^c_j; {\vect b}^s_j \circ {\vect b}^c_j;\sum_{i=1}^{m} \beta_{ij} {\vect r}_{ji}]) \,,
\end{align}
}
\noindent where a heuristic matching trick with difference and element-wise product is used~\citep{DBLP:conf/acl/MouMLX0YJ16,DBLP:conf/acl/ChenZLWJI17}. 
The last terms in Equation (\ref{eq:infer1})(\ref{eq:infer2}) are used to obtain word-level inference information from external knowledge. Take Equation (\ref{eq:infer1}) as example, ${\vect r}_{ij}$ is the relation feature between the $i$-th word in the premise and the $j$-th word in the hypothesis, but we care more about semantic relation between aligned word pairs between the premise and the hypothesis. Thus, we use a soft-aligned version through the soft-alignment weight $\alpha_{ij}$. For the $i$-th word in the premise, the last term in Equation (\ref{eq:infer1}) is a word-level inference information based on external knowledge between the $i$-th word and the aligned word. The same calculation for hypothesis is performed in Equation (\ref{eq:infer2}). $G$ is a non-linear mapping function to reduce dimensionality. Specifically, we use a 1-layer feed-forward neural network with the ReLU activation function with a shortcut connection, i.e., concatenate the hidden states after ReLU with the input $\sum_{j=1}^{n} \alpha_{ij} {\vect r}_{ij}$ (or $\sum_{i=1}^{m} \beta_{ij} {\vect r}_{ji}$) as the output ${\vect a}^m_i$ (or ${\vect b}^m_j$).

\subsection{Knowledge-Enhanced Inference Composition}
In this component, we introduce knowledge-enriched inference composition.
To determine the overall inference relationship between the premise and the hypothesis, we need to explore a composition layer to compose the local inference vectors (${\vect a}^m$ and ${\vect b}^m$) collected above:
\begin{align}
{\vect a}^v_i &= \mathrm{Composition}({\vect a}^m,i) \,, \\
{\vect b}^v_j &= \mathrm{Composition}({\vect b}^m,j) \,.
\end{align}
Here, we also use BiLSTMs as building blocks for the composition layer, but the responsibility of BiLSTMs in the inference composition layer is completely different from that in the input encoding layer. The BiLSTMs here read local inference vectors (${\vect a}^m$ and ${\vect b}^m$) and learn to judge the types of local inference relationship and distinguish crucial local inference vectors for overall sentence-level inference relationship. Intuitively, the final prediction is likely to depend on word pairs appearing in external knowledge that have some semantic relation.
Our inference model converts the output hidden vectors of BiLSTMs to the fixed-length vector with pooling operations and puts it into the final classifier to determine the overall inference class. Particularly, in addition to using mean pooling and max pooling similarly to ESIM~\citep{DBLP:conf/acl/ChenZLWJI17}, we propose to use weighted pooling based on external knowledge to obtain a fixed-length vector as in Equation (\ref{eq:weight:1})(\ref{eq:weight:2}).  
{\fontsize{10pt}{1.0cm}
\begin{align}
\label{eq:weight:1}
{\vect a}^{\mathrm{w}} &= \sum_{i=1}^m \frac{\exp(H(\sum_{j=1}^n \alpha_{ij}{\vect r}_{ij}))}{\sum_{i=1}^m\exp(H(\sum_{j=1}^n \alpha_{ij}{\vect r}_{ij}))} {\vect a}^v_i \,,\\
\label{eq:weight:2}
{\vect b}^{\mathrm{w}} &= \sum_{j=1}^n \frac{\exp(H(\sum_{i=1}^m \beta_{ij} {\vect r}_{ji}))}{\sum_{j=1}^n\exp(H(\sum_{i=1}^m \beta_{ij}{\vect r}_{ji}))} {\vect b}^v_j \,. 
\end{align}
}
In our experiments, we regard the function $H$ as a 1-layer feed-forward neural network with ReLU activation function. We concatenate all pooling vectors, i.e., mean, max, and weighted pooling, into the fixed-length vector and then put the vector into the final multilayer perceptron (MLP) classifier. The MLP has one hidden layer with \textit{tanh} activation and \textit{softmax} output layer in our experiments. The entire model is trained end-to-end, through minimizing the cross-entropy loss. 

\section{Experiment Set-Up}
\label{sec:setup}
\subsection{Representation of External Knowledge}
\paragraph{Lexical Semantic Relations} As described in Section \ref{sec:knowledge}, to incorporate external knowledge (as a knowledge vector ${\vect r}_{ij}$) to the state-of-the-art neural network-based NLI models, we first explore semantic relations in WordNet~\citep{DBLP:journals/cacm/Miller95}, motivated by~\citet{MacCartneyThesis}. Specifically, the relations of lexical pairs are derived as described in~\ref{itemB}-\ref{itemE} below. Instead of using Jiang-Conrath WordNet distance metric~\citep{DBLP:conf/rocling/JiangC97}, which does not improve the performance of our models on the development sets, we add a new feature, i.e., \textit{co-hyponyms}, which consistently benefit  our models. 

\begin{enumerate}[(1)]
\itemsep 0em
\item \label{itemB} \textit{Synonymy}: It takes the value $1$ if the words in the pair are synonyms in WordNet (i.e., belong to the same synset), and $0$ otherwise. For example, [felicitous, good] = $1$, [dog, wolf] = $0$. 
\item \textit{Antonymy}: It takes the value $1$ if the words in the pair are antonyms in WordNet, and $0$ otherwise. For example, [wet, dry] = $1$. 
\item \textit{Hypernymy}: It takes the value $1-n/8$ if one word is a (direct or indirect) hypernym of the other word in WordNet, where $n$ is the number of edges between the two words in hierarchies, and $0$ otherwise. Note that we ignore pairs in the hierarchy which have more than 8 edges in between. For example, [dog, canid] = $0.875$, [wolf, canid] = $0.875$, [dog, carnivore] = $0.75$, [canid, dog] = $0$
\item \label{itemE} \textit{Hyponymy}: It is simply the inverse of the hypernymy feature.
For example, [canid, dog] = $0.875$, [dog, canid] = $0$.
\item  \textit{Co-hyponyms}: It takes the value $1$ if the two words have the same hypernym but they do not belong to the same synset, and $0$ otherwise. For example, [dog, wolf] = $1$.
\end{enumerate}

As discussed above, we expect features like \textit{synonymy}, \textit{antonymy}, \textit{hypernymy}, \textit{hyponymy} and \textit{co-hyponyms} would help model co-attention alignment between the premise and the hypothesis. Knowledge of \textit{hypernymy} and \textit{hyponymy} may help capture entailment; knowledge of \textit{antonymy} and \textit{co-hyponyms} may help model contradiction. Their final contributions will be learned in end-to-end model training. We regard the vector ${\vect r}\in \RR^{d_r}$ as the relation feature derived from external knowledge, where $d_r$ is $5$ here. In addition, Table~\ref{tab:stat} reports some key statistics of these features.

\begin{table}[ht]
\renewcommand{\arraystretch}{0.9}
\begin{center}
\scalebox{0.9}{
\begin{tabular}{l r r}
\toprule
\multicolumn{1}{l}{\textbf{Feature}} & \multicolumn{1}{r}{\textbf{\#Words}} & \multicolumn{1}{r}{\textbf{\#Pairs}}  \\
\midrule
\textit{Synonymy} &84,487  &237,937\\
\textit{Antonymy} &6,161 &6,617\\
\textit{Hypernymy} &57,475 &753,086 \\
\textit{Hyponymy} &57,475 &753,086 \\
\textit{Co-hyponyms} & 53,281 & 3,674,700 \\
\bottomrule
\end{tabular}
}
\end{center}
\caption{Statistics of lexical relation features.}
\label{tab:stat}
\end{table}

In addition to the above relations, we also use more relation features in WordNet, including \textit{instance}, \textit{instance of}, \textit{same instance}, \textit{entailment}, \textit{member meronym}, \textit{member holonym}, \textit{substance meronym}, \textit{substance holonym}, \textit{part meronym}, \textit{part holonym}, summing up to 15 features, but these additional features do not bring further improvement on the development dataset, as also discussed in Section ~\ref{sec:res}. 

\paragraph{Relation Embeddings} In the most recent years graph embedding has been widely employed to learn representation for vertexes and their relations in a graph. In our work here, we also capture the relation between any two words in WordNet through relation embedding. Specifically, we employed TransE~\citep{DBLP:conf/nips/BordesUGWY13}, a widely used graph embedding methods, to capture relation embedding between any two words. We used two typical approaches to obtaining the relation embedding. The first directly uses 18 relation embeddings pretrained on the WN18 dataset~\citep{DBLP:conf/nips/BordesUGWY13}. Specifically, if a word pair has a certain type relation, we take the corresponding relation embedding. Sometimes, if a word pair has multiple relations among the 18 types; we take an average of the relation embedding. The second approach uses TransE's word embedding (trained on WordNet) to obtain relation embedding, through the objective function used in TransE, i.e., ${\vect l} \approx {\vect t} - {\vect h}$, where ${\vect l}$ indicates relation embedding, ${\vect t}$ indicates tail entity embedding, and ${\vect h}$ indicates head entity embedding. 

Note that in addition to relation embedding trained on WordNet, other relational embedding resources exist; e.g., that trained on Freebase (WikiData)~\citep{DBLP:conf/aaai/BollackerCT07}, but such knowledge resources are mainly about facts (e.g., relationship between Bill Gates and Microsoft) and are less for commonsense knowledge used in general natural language inference (e.g., the color yellow potentially contradicts red).

\subsection{NLI Datasets}
In our experiments, we use Stanford Natural Language Inference (SNLI) dataset~\citep{DBLP:conf/emnlp/BowmanAPM15} and Multi-Genre Natural Language Inference (MultiNLI)~\citep{DBLP:journals/corr/WilliamsNB17} dataset, which focus on three basic relations between a premise and a potential hypothesis: the premise entails the hypothesis (\textit{entailment}), they contradict each other (\textit{contradiction}), or they are not related (\textit{neutral}). 
We use the same data split as in previous work ~\citep{DBLP:conf/emnlp/BowmanAPM15,DBLP:journals/corr/WilliamsNB17} and classification accuracy as the evaluation metric. 
In addition, we test our models (trained on the SNLI training set) on a new test set~\citep{glockner_acl18}, which assesses the lexical inference abilities of NLI systems and consists of 8,193 samples.
WordNet 3.0~\citep{DBLP:journals/cacm/Miller95} is used to extract semantic relation features between words. The words are lemmatized using Stanford CoreNLP 3.7.0~\citep{DBLP:conf/acl/ManningSBFBM14}. The premise and the hypothesis sentences fed into the input encoding layer are tokenized.

\subsection{Training Details}
For duplicability, we release our code\footnote{https://github.com/lukecq1231/kim}. All our models were strictly selected on the development set of the SNLI data and the in-domain development set of MultiNLI and were then tested on the corresponding test set. The main training details are as follows: the dimension of the hidden states of LSTMs and word embeddings are $300$. The word embeddings are initialized by \textit{300D GloVe 840B}~\citep{DBLP:conf/emnlp/PenningtonSM14}, and out-of-vocabulary words among them are initialized randomly. All word embeddings are updated during training. Adam~\citep{DBLP:journals/corr/KingmaB14} is used for optimization with an initial learning rate of $0.0004$. The mini-batch size is set to $32$. Note that the above hyperparameter settings are same as those used in the baseline ESIM~\citep{DBLP:conf/acl/ChenZLWJI17} model. ESIM is a strong NLI baseline framework with the source code made available at \textit{https://github.com/lukecq1231/nli} (the ESIM core code has also been adapted to summarization~\citep{DBLP:conf/ijcai/ChenZLWJ16} and question-answering tasks~\citep{Zhang:qa:2017}). 

The trade-off $\lambda$ for calculating co-attention in Equation~(\ref{eq:eij}) is selected in $[0.1, 0.2, 0.5, 1, 2, 5, 10, 20, 50]$ based on the development set. When training TransE for WordNet, relations are represented with vectors of $20$ dimension.

\section{Experimental Results}
\label{sec:res}
\subsection{Overall Performance} 

Table~\ref{tab:snli} shows the results of state-of-the-art models on the SNLI dataset. 
Among them, ESIM~\citep{DBLP:conf/acl/ChenZLWJI17} is one of the previous state-of-the-art systems with an 88.0\% test-set accuracy. The proposed model, namely Knowledge-based Inference Model (\textbf{KIM}), which enriches ESIM with external knowledge, obtains an accuracy of 88.6\%, the best single-model performance reported on the SNLI dataset. The difference between ESIM and KIM is statistically significant under the one-tailed paired $t$-test at the 99\% significance level. Note that the KIM model reported here uses five semantic relations described in Section~\ref{sec:setup}. In addition to that, we also use 15 semantic relation features, which does not bring additional gains in performance. These results highlight the effectiveness of the five semantic relations described in Section~\ref{sec:setup}. To further investigate external knowledge, we add TransE relation embedding, and again no further improvement is observed on both the development and test sets when TransE relation embedding is used (concatenated) with the semantic relation vectors. We consider this is due to the fact that TransE embedding is not specifically sensitive to inference information; e.g., it does not model co-hyponyms features, and its potential benefit has already been covered by the semantic relation features used.   

\begin{table}[t!]
\renewcommand{\arraystretch}{0.9}
\begin{center}
\scalebox{0.9}{
\begin{tabular}{l r}
\toprule
\multicolumn{1}{l}{\textbf{Model}} & \multicolumn{1}{r}{\textbf{Test}}\\
\midrule
LSTM Att.~\citep{DBLP:journals/corr/RocktaschelGHKB15}  & 83.5 \\
DF-LSTMs~\citep{DBLP:conf/acl/LiuQCH16} & 84.6 \\
TC-LSTMs~\citep{DBLP:conf/emnlp/LiuQZCH16} & 85.1 \\
Match-LSTM~\citep{DBLP:conf/naacl/WangJ16} & 86.1 \\       
LSTMN~\citep{DBLP:conf/emnlp/0001DL16} & 86.3 \\
Decomposable Att.~\citep{DBLP:conf/emnlp/ParikhT0U16} & 86.8 \\
NTI~\citep{DBLP:conf/eacl/YuM17} & 87.3 \\
Re-read LSTM~\citep{DBLP:conf/coling/ShaCSL16} &  87.5 \\
BiMPM~\citep{DBLP:conf/ijcai/WangHF17} &  87.5 \\
DIIN~\citep{DBLP:journals/corr/abs-1709-04348} & 88.0 \\
BCN + CoVe~\citep{DBLP:conf/nips/McCannBXS17} & 88.1 \\
CAFE~\citep{DBLP:journals/corr/abs-1801-00102} & 88.5 \\
\midrule
ESIM~\citep{DBLP:conf/acl/ChenZLWJI17} &	88.0 \\
KIM (This paper) & \textbf{88.6} \\
\bottomrule
\end{tabular}
}
\end{center}
\caption{Accuracies of models on SNLI. }
\label{tab:snli}
\end{table}

Table~\ref{tab:multinli} shows the performance of models on the MultiNLI dataset. The baseline ESIM achieves 76.8\% and 75.8\% on in-domain and cross-domain test set, respectively. If we extend the ESIM with external knowledge, we achieve significant gains to 77.2\% and 76.4\% respectively. Again, the gains are consistent on SNLI and MultiNLI, and we expect they would be orthogonal to other factors when external knowledge is added into other state-of-the-art models.

\begin{table}[t!]
\renewcommand{\arraystretch}{0.9}
\centering
\scalebox{0.9}{
\begin{tabular}{l r r}
\toprule
\multicolumn{1}{l}{\textbf{Model}} & \multicolumn{1}{r}{\textbf{In}} & \multicolumn{1}{r}{\textbf{Cross}}\\
\midrule
CBOW~\citep{DBLP:journals/corr/WilliamsNB17} & 64.8 & 64.5 \\
BiLSTM~\citep{DBLP:journals/corr/WilliamsNB17} &  66.9 & 66.9\\
DiSAN~\citep{DBLP:journals/corr/abs-1709-04696} & 71.0 & 71.4 \\
Gated BiLSTM~\citep{DBLP:conf/repeval/ChenZLWJI17} & 73.5 & 73.6\\
SS BiLSTM~\citep{DBLP:conf/repeval/NieB17} & 74.6 & 73.6 \\
DIIN *~\citep{DBLP:journals/corr/abs-1709-04348} & 77.8 & \textbf{78.8} \\
CAFE~\citep{DBLP:journals/corr/abs-1801-00102} & \textbf{78.7} & 77.9 \\
\midrule
ESIM~\citep{DBLP:conf/acl/ChenZLWJI17} & 76.8 & 75.8 \\
KIM (This paper) & \textbf{77.2} & \textbf{76.4} \\
\bottomrule
\end{tabular}
}
\caption{Accuracies of models on MultiNLI. * indicates models using extra SNLI training set.}
\label{tab:multinli}
\end{table}

\subsection{Ablation Results}
\begin{figure}[!htb]
	\centering
	\includegraphics[width=0.8\linewidth]{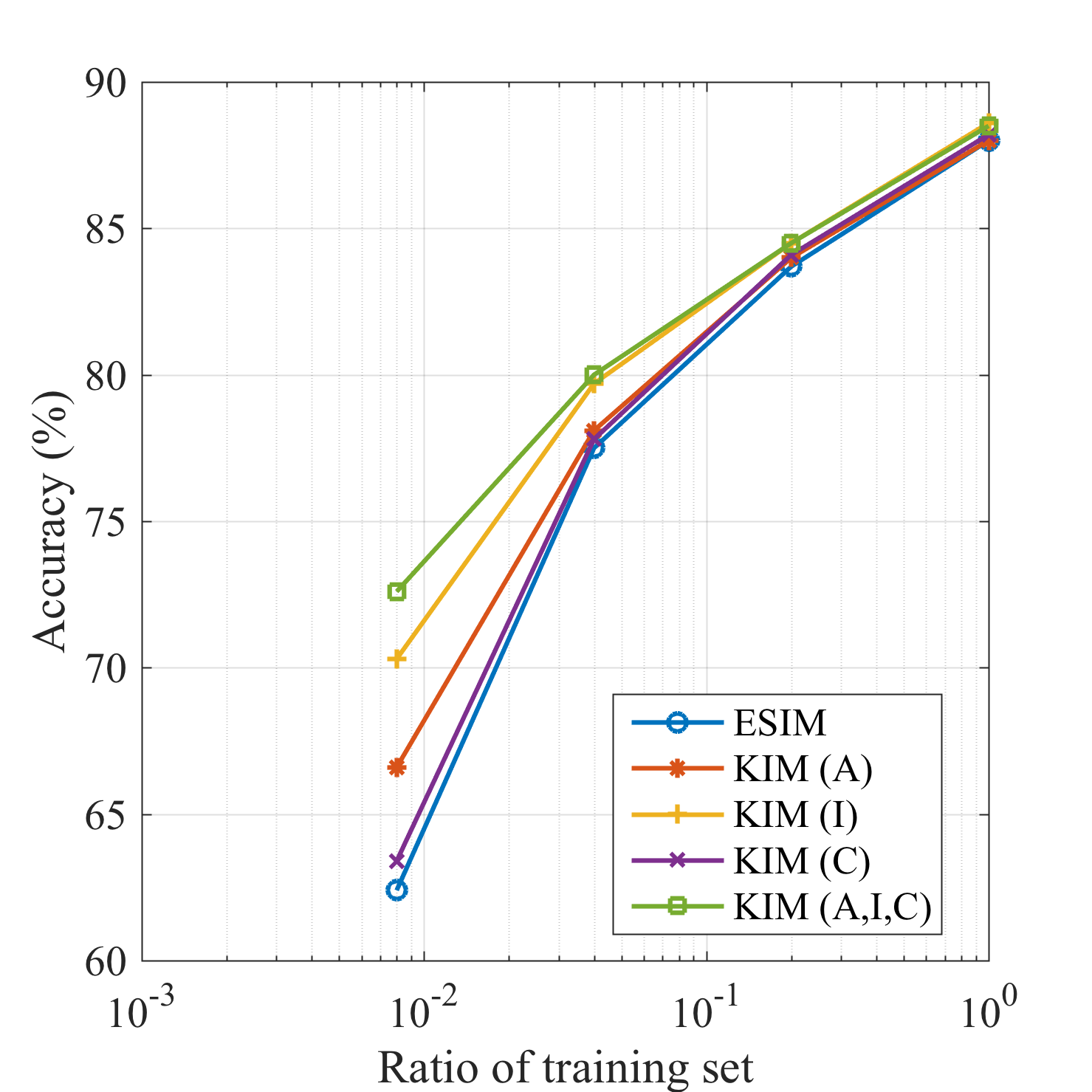}
	\caption{Accuracies of models of incorporating external knowledge into different NLI components, under different sizes of training data (0.8\%, 4\%, 20\%, and the entire training data).}
	\label{fig:curve1}
\end{figure}

Figure~\ref{fig:curve1} displays the ablation analysis of different components when using the external knowledge. To compare the effects of external knowledge under different training data scales, we randomly sample different ratios of the entire training set, i.e., 0.8\%, 4\%, 20\% and 100\%. ``A'' indicates adding external knowledge in calculating the co-attention matrix as in Equation~(\ref{eq:eij}), ``I'' indicates adding external knowledge in collecting local inference information as in Equation~(\ref{eq:infer1})(\ref{eq:infer2}), and ``C'' indicates adding external knowledge in composing inference as in Equation~(\ref{eq:weight:1})(\ref{eq:weight:2}). When we only have restricted training data, i.e., 0.8\% training set (about 4,000 samples), the baseline ESIM has a poor accuracy of 62.4\%. When we only add external knowledge in calculating co-attention (``A''), the accuracy increases to 66.6\% (+ absolute 4.2\%). When we only utilize external knowledge in collecting local inference information (``I''), the accuracy has a significant gain, to 70.3\% (+ absolute 7.9\%). When we only add external knowledge in inference composition (``C''), the accuracy gets a smaller gain to 63.4\% (+ absolute 1.0\%). The comparison indicates that ``I'' plays the most important role among the three components in using external knowledge. Moreover, when we compose the three components (``A,I,C''), we obtain the best result of 72.6\% (+ absolute 10.2\%). When we use more training data, i.e., 4\%, 20\%, 100\% of the training set, only ``I'' achieves a significant gain, but ``A'' or ``C'' does not bring any significant improvement. The results indicate that external semantic knowledge only helps co-attention and composition when limited training data is limited, but always helps in collecting local inference information. Meanwhile, for less training data, $\lambda$ is usually set to a larger value. For example, the optimal $\lambda$ on the development set is $20$ for 0.8\% training set, $2$ for the 4\% training set, $1$ for the 20\% training set and $0.2$ for the 100\% training set.

\begin{figure}[!htb]
	\centering
	\includegraphics[width=0.8\linewidth]{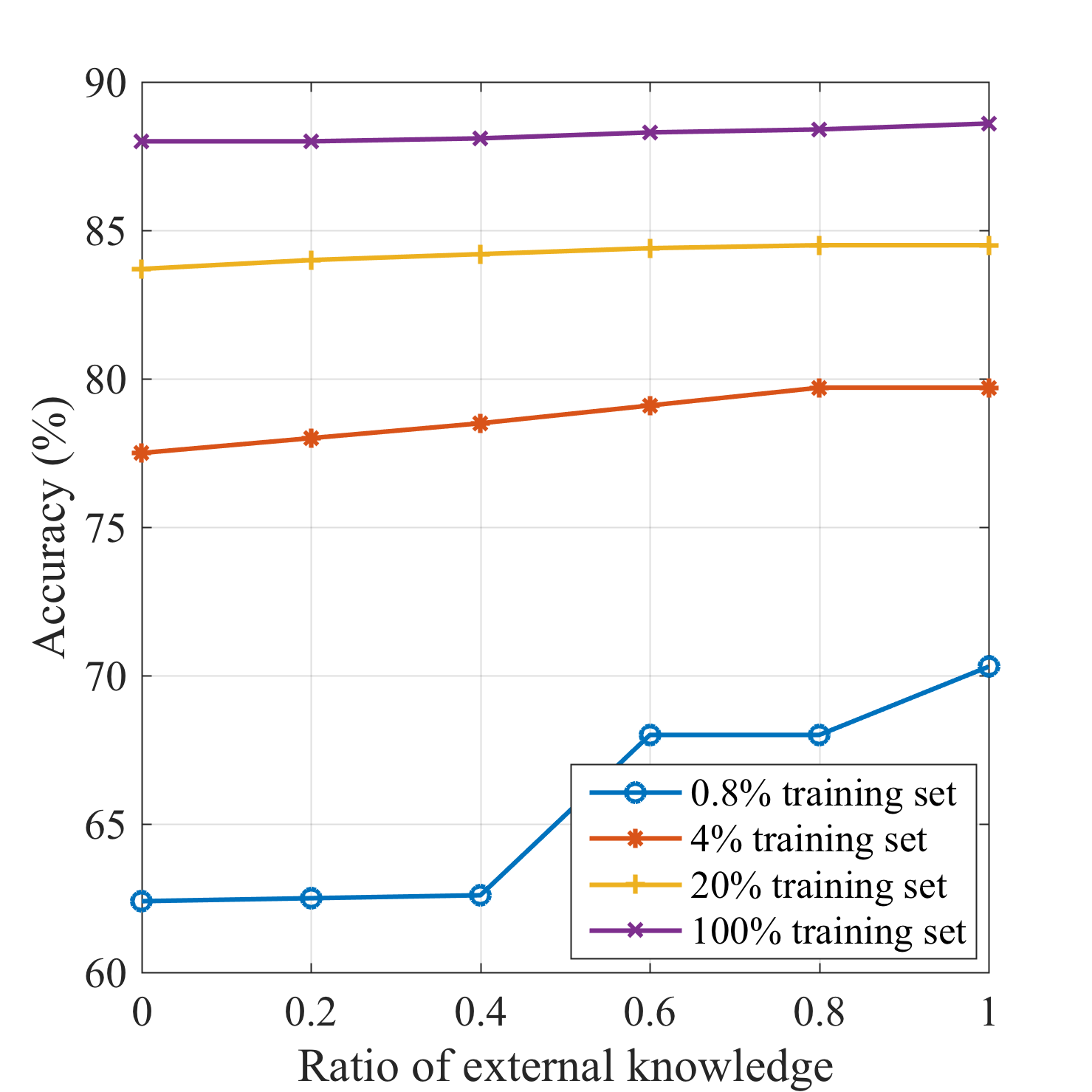}
	\caption{Accuracies of models under different sizes of external knowledge. More external knowledge corresponds to higher accuracies.}
	\label{fig:curve2}
\end{figure}

Figure~\ref{fig:curve2} displays the results of using different ratios of external knowledge (randomly keep different percentages of whole lexical semantic relations) under different sizes of training data. Note that here we only use external knowledge in collecting local inference information as it always works well for different scale of the training set. Better accuracies are achieved when using more external knowledge. Especially under the condition of restricted training data (0.8\%), the model obtains a large gain when using more than half of external knowledge.

\subsection{Analysis on the ~\citep{glockner_acl18} Test Set}
In addition, Table \ref{tab:newtest} shows the results on a newly published test set~\citep{glockner_acl18}. Compared with the performance on the SNLI test set, the performance of the three baseline models dropped substantially on the~\citep{glockner_acl18} test set, with the differences ranging from 22.3\% to 32.8\% in accuracy. Instead, the proposed KIM achieves 83.5\% on this test set (with only a 5.1\% drop in performance), which demonstrates its better ability of utilizing lexical level inference and hence better generalizability.

Figure \ref{tab:category} displays the accuracy of ESIM and KIM in each replacement-word category of  the~\citep{glockner_acl18} test set. KIM outperforms ESIM in 13 out of 14 categories, and only performs worse on synonyms. 

\begin{table}[t!]
\renewcommand{\arraystretch}{0.9}
\begin{center}
\scalebox{0.9}{
\begin{tabular}{l r r}
\toprule
\multicolumn{1}{l}{\textbf{Model}} & \multicolumn{1}{r}{\textbf{SNLI}} & \multicolumn{1}{r}{\textbf{Glockner's($\Delta$)}}\\
\midrule
\citep{DBLP:conf/emnlp/ParikhT0U16}* & 84.7 & 51.9 (-32.8)\\
\citep{DBLP:conf/repeval/NieB17}* & 86.0 & 62.2 (-23.8) \\
ESIM * & 87.9 & 65.6 (-22.3)\\
KIM (This paper)  & \textbf{88.6} & \textbf{83.5} (~~-5.1)\\
\bottomrule
\end{tabular}
}
\end{center}
\caption{Accuracies of models on the SNLI and ~\citep{glockner_acl18} test set. * indicates the results taken from \citep{glockner_acl18}.}
\label{tab:newtest}
\end{table}

\begin{table}[t!]
\renewcommand{\arraystretch}{0.9}
\centering
\scalebox{0.9}{
\begin{tabular}{l r r r}
\toprule
\multicolumn{1}{l}{\textbf{Category}} & \multicolumn{1}{l}{\textbf{Instance}} & \multicolumn{1}{r}{\textbf{ESIM}} & \multicolumn{1}{r}{\textbf{KIM}} \\
\midrule
Antonyms & 1,147 & 70.4 &  \textbf{86.5}\\
Cardinals & 759 & 75.5 & \textbf{93.4}\\
Nationalities & 755 & 35.9 & \textbf{73.5}\\
Drinks & 731 & 63.7 & \textbf{96.6}\\
Antonyms WordNet & 706 & 74.6 & \textbf{78.8}\\
Colors & 699 & 96.1 & \textbf{98.3}\\
Ordinals & 663 & 21.0 & \textbf{56.6}\\
Countries & 613 & 25.4 & \textbf{70.8}\\
Rooms & 595 & 69.4 &\textbf{ 77.6}\\
Materials & 397 & 89.7 & \textbf{98.7}\\
Vegetables & 109 & 31.2 & \textbf{79.8}\\
Instruments & 65 & 90.8 & \textbf{96.9}\\
Planets & 60 & 3.3 & \textbf{5.0}\\
Synonyms & 894 & \textbf{99.7} & 92.1\\ 
\midrule
Overall & 8,193 & 65.6 & \textbf{83.5} \\
\bottomrule
\end{tabular}
}
\caption{The number of instances and accuracy per category achieved by ESIM and KIM on the~\citep{glockner_acl18} test set.}
\label{tab:category}
\end{table}

\subsection{Analysis by Inference Categories}

We perform more analysis (Table~\ref{tab:error}) using the supplementary annotations provided by the MultiNLI dataset~\citep{DBLP:journals/corr/WilliamsNB17}, which have 495 samples (about 1/20 of the entire development set) for both in-domain and out-domain set. We compare against the model outputs of the ESIM model across 13 categories of inference. Table~\ref{tab:error} reports the results. We can see that KIM outperforms ESIM on overall accuracies on both in-domain and cross-domain subset of development set. KIM outperforms or equals ESIM in 10 out of 13 categories on the cross-domain setting, while only 7 out of 13 categories on in-domain setting. It indicates that external knowledge helps more in cross-domain setting. Especially, for antonym category in cross-domain set, KIM outperform ESIM significantly (+ absolute 5.0\%) as expected, because antonym feature captured by external knowledge would help unseen cross-domain samples.

\begin{table}[t!]
\renewcommand{\arraystretch}{0.9}
\centering
\scalebox{0.9}{
\begin{tabular}{l r r r r}
\toprule
\multicolumn{1}{l}{\textbf{Category}}& \multicolumn{2}{r}{\textbf{In-domain}} & \multicolumn{2}{r}{\textbf{Cross-domain}}\\
 & \multicolumn{1}{r}{{ESIM}} & \multicolumn{1}{r}{{KIM}} & \multicolumn{1}{r}{{ESIM}} & \multicolumn{1}{r}{{KIM}}\\
\midrule
Active/Passive & \textbf{93.3} & \textbf{93.3} & \textbf{100.0} & \textbf{100.0} \\
Antonym	& \textbf{76.5} & \textbf{76.5} & 70.0 & \textbf{75.0}\\
Belief & 72.7 & \textbf{75.8} & 75.9 & \textbf{79.3} \\
Conditional	& \textbf{65.2} & \textbf{65.2} & 61.5 & \textbf{69.2} \\
Coreference & \textbf{80.0} & 76.7 &  \textbf{75.9} & \textbf{75.9} \\
Long sentence & \textbf{82.8} &78.8 & 69.7 &\textbf{73.4}\\
Modal & \textbf{80.6} & 79.9 & 77.0 & \textbf{80.2} \\
Negation &	76.7 & \textbf{79.8}& \textbf{73.1} & 71.2 \\
Paraphrase & \textbf{84.0} & 72.0 & 86.5& \textbf{89.2}\\
Quantity/Time & \textbf{66.7} & \textbf{66.7} & 56.4 & \textbf{59.0} \\
Quantifier & \textbf{79.2} & 78.4 & 73.6  & \textbf{77.1}\\
Tense & 74.5 & \textbf{78.4} & \textbf{72.2} &66.7\\
Word overlap & \textbf{89.3} & 85.7 & \textbf{83.8} & 81.1 \\
\midrule
Overall &77.1 & \textbf{77.9}& 76.7 & \textbf{77.4} \\
\bottomrule
\end{tabular}
}
\caption{Detailed Analysis on MultiNLI. }
\label{tab:error}
\end{table}

\subsection{Case Study}
Table~\ref{tab:examples} includes some examples from the SNLI test set, where KIM successfully predicts the inference relation and ESIM fails. In the first example, the premise is ``An African person standing in a \textbf{wheat} field'' and the hypothesis ``A person standing in a \textbf{corn} field''. As the KIM model knows that ``wheat'' and ``corn'' are both a kind of cereal, i.e, the \textit{co-hyponyms} relationship in our relation features, KIM therefore predicts the premise contradicts the  hypothesis. However, the baseline ESIM cannot learn the relationship between ``wheat'' and ``corn'' effectively due to lack of enough samples in the training sets. With the help of external knowledge, i.e., ``wheat'' and ``corn'' having the same hypernym ``cereal'', KIM predicts contradiction correctly.

\begin{table}[t!]
\renewcommand{\arraystretch}{0.9}
\centering
\scalebox{0.9}{
\begin{tabular}{c p{6.8cm}}
\toprule
\textbf{P/G} & \textbf{Sentences}\\
\midrule
\textit{e/c}	 & \textit{p}: An African person standing in a \textbf{wheat} field. \\
& \textit{h}: A person standing in a \textbf{corn} field. \\
\midrule
\textit{e/c} &\textit{p}: Little girl is flipping an \textbf{omelet} in the kitchen. \\
& \textit{h}: A young girl cooks \textbf{pancakes}. \\
\midrule
\textit{c/e}	&\textit{p}: A middle eastern \textbf{marketplace}. \\
& \textit{h}: A middle easten \textbf{store}. \\
\midrule
\textit{c/e}	&\textit{p}: Two boys are swimming with boogie \textbf{boards}. \\
& \textit{h}: Two boys are swimming with their \textbf{floats}. \\
\bottomrule
\label{tab:examples}
\end{tabular}
}
\caption{Examples. Word in bold are key words in making final prediction. \textbf{P} indicates a predicted label and \textbf{G} indicates gold-standard label. \textit{e} and \textit{c} denote \textit{entailment} and \textit{contradiction}, respectively.}
\label{tab:examples}
\end{table}

\section{Conclusions}
Our neural-network-based model for natural language inference with external knowledge, namely KIM, achieves the state-of-the-art accuracies. The model is equipped with external knowledge in its main components, specifically, in calculating co-attention, collecting local inference, and composing inference. We provide detailed analyses on our model and results. The proposed model of infusing neural networks with external knowledge may also help shed some light on tasks other than NLI.

\section*{Acknowledgments}
We thank Yibo Sun and Bing Qin for early helpful discussion.

\newpage
\bibliography{acl2018}
\bibliographystyle{acl_natbib}

\end{document}